\pdfoutput=1
\documentclass[letterpaper,11pt]{article}

\title
{
	``On Event Structure in the Torn Dress''
	\footnote
	{
		The title is comprised from part of titles of major
		papers being discussed, namely Pustejovsky \cite{pustejovsky}
		and Fong \cite{fongdress}.
	}
}

{\author
	{
		\vspace{150pt}\\\hline\\
		\textbf{\Large Serguei A. Mokhov}\\
		email: \texttt{mokhov@cs.concordia.ca}\\
		\vspace{40pt}
	}
}

\date
{
	$Revision: 1.11 $ \\ Date: Mon Nov 17 23:40:54 EST 2003
\\
	\vspace{40pt}
	Montr\'eal, Qu\'ebec\\
	Canada
}
\usepackage{graphicx}
\usepackage{latexsym}
\usepackage{makeidx}

\makeindex

\newcommand{\xf}[1]{Figure~\ref{#1}}

\newcommand{\xs}[1]{Section~\ref{#1}}

\newcommand{\tab}[1]{\hspace{#1pt}}

\newcommand{\shrule}[0]{\vspace{3pt}\hrule\vspace{6pt}}
\newcommand{\ehrule}[0]{\vspace{6pt}\hrule\vspace{3pt}}

\begin{document}

\begin{titlepage}
	\maketitle
	\thispagestyle{empty}
\end{titlepage}

\clearpage

\pagenumbering{roman}
\tableofcontents
\clearpage
\pagenumbering{arabic}

\clearpage

\section{Introduction}

This document discusses relevance of the techniques employed
in \cite{pustejovsky} to example sentences and ideas presented
in \cite{fongdress}. The document goes beyond just a mechanical
attempt to apply the techniques from one paper to another; it
also discusses the pros, cons, and other findings in the two
papers. A relationship is made to other works in Lexical Semantics
and Lexicon; the sources are cited where appropriate. If there
are no citations, the text is a sole responsibility of the author.

\section{Goal}

The main goal of this work is to illustrate on how to one can
possibly represent semantics of sentences using tools presented
in the papers (\cite{pustejovsky}, \cite{fongdress}) and others
and derive a computational lexicon.

\section{Semantic Analysis}\label{sect:semanalysis}

The below sentence samples have been taken from \cite{fongdress}.
All the follow up discussion will be around these examples.

\begin{enumerate}

%1
\item
	\begin{enumerate}
	\item
	{\bf Cathie mended the {\it torn} dress.}
	\item
	{\bf Cathie mended the {\it red} dress.}
	\end{enumerate}

	As Fong herself says in \cite{fongdress}, the both events of the above
	fit the same event template (LCS):
	
	(c) [$x$ CAUSE [BECOME [$y$ $<$mended$>$]]]
	
	This is possible for the lexeme mend$^1$ because
	1b has implicit {\it torn} adjective in it. I claim that
	(b) and the below are equivalent:
	
	(d) Cathie mended the {\it torn red} dress.
	
	The {\it red} is just a bit more specific type of dress
	being mended. Otherwise, why would one mend a dress
	which is not torn? The {\it torn} state of the dress 
	changes to become {\it mended}; hence, state transition is
	in place. Another part of dress' state, as being {\it red}
	is undefined (but is likely to be still the same) at the 
	event culmination.
	
	We can build the event structure, ES, LCS', and LCS following
	the style of Pustejovsky in \cite{pustejovsky} and Fong in \cite{fongdress}.
	I have replaced the {\it not} notation to ``!''. For ES and LCS' I use
	the notation from Pustejovsky, and LCS is from Fong because it resembles
	more conventional use of predicates than that of Pustejovsky.
	
	\clearpage
	
	{\shrule}
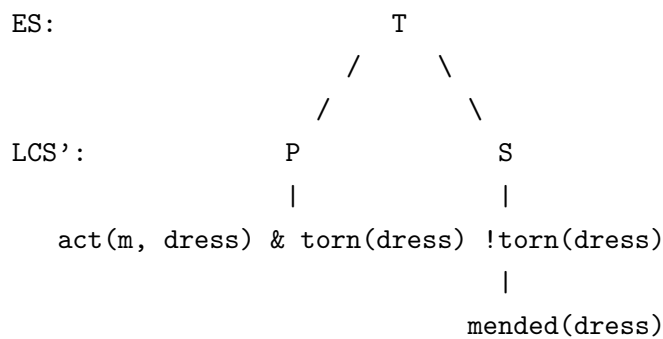
	\begin{verbatim}
	ES:                      T
	                      /     \
	                    /         \
	LCS':             P             S
	                  |             |
	   act(m, dress) & torn(dress) !torn(dress)
	                                |
	                              mended(dress)
	  
	LCS: [m CAUSE [BECOME [dress <mended>]]]
	\end{verbatim}
	{\ehrule}

	As for the refined structure for a specific, {\it red},
	dress it won't change much, that just a part of state
	will remain the same.
	
	{\shrule}
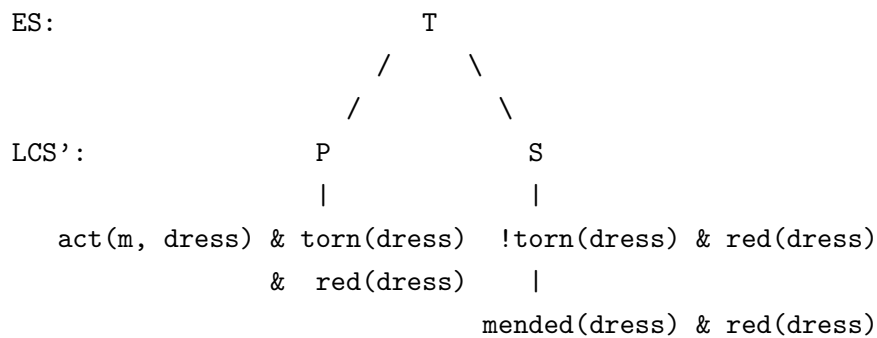
	\begin{verbatim}
	ES:                        T
	                        /     \
	                      /         \
	LCS':               P             S
	                    |             |
	   act(m, dress) & torn(dress)  !torn(dress) & red(dress)
	                 &  red(dress)    |
	                               mended(dress) & red(dress)
	  
	LCS: [m CAUSE [BECOME [dress <mended>]]] & [IS [dress <red>]]
	\end{verbatim}
	{\ehrule}

	One may argue, that the dress might change the color
	during the mending process because we can only guess
	that it is still red but don't have enough evidence
	from the sentence being analyzed that the red property
	persists. Then we could adjust the state template
	\verb+[IS [dress <red>]]+ to be 
	\verb+[IS [dress <red>] | IS [dress [NOT <red>]]]+
	and assign the first part of the disjunct a higher
	likelihood than the second (I use ``\verb+|+'' to indicate OR
	and ``\&'' to indicate AND).

	In this analysis if we assume {\it red} being a constant
	throughout the process we can claim than the original
	template (c) holds. Since it is a transition, we could
	also write (notational sugar):
	 
	(e) $T_{torn}^{mended}$(dress) \& red(dress).

\item
	\begin{enumerate}
	\item
	{\bf The plumber fixed every {\it leaky} faucet.}
	\item
	{\bf The plumber fixed every {\it blue} faucet.}
	\end{enumerate}
	
	These two examples are similar to that of (1).
	Both assume the faucets fixed were leaking water,
	hence there's a state transition with respect
	to the property of every fixed faucet to become
	not leaky. Additional semantic bit in here is that
	the adjective {\it every} makes it plural and this
	fact may shift the event type sometimes.
	
	{\shrule}
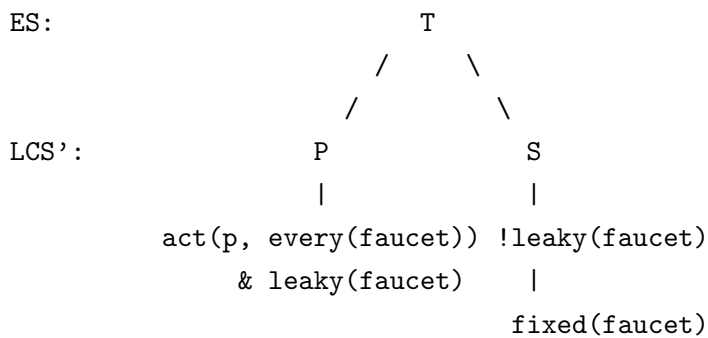
	\begin{verbatim}
	ES:                        T
	                        /     \
	                      /         \
	LCS':               P             S
	                    |             |
	          act(p, every(faucet)) !leaky(faucet)
	               & leaky(faucet)    |
	                                 fixed(faucet)
	  
	LCS (1): [p CAUSE [BECOME [EVERY [faucet <fixed>]]]]
	LCS (2): [p CAUSE [BECOME [faucets <fixed>]]]
	\end{verbatim}
	{\ehrule}

	The second case restricts the act of
	fixing only to leaky blue faucets (there may be, let say,
	leaky red faucets that have not been not fixed).
	The {\it blue} property of (b) does
	not change since it is not in the opposition
	to or in a synnet with {\it leaky}.

	{\shrule}
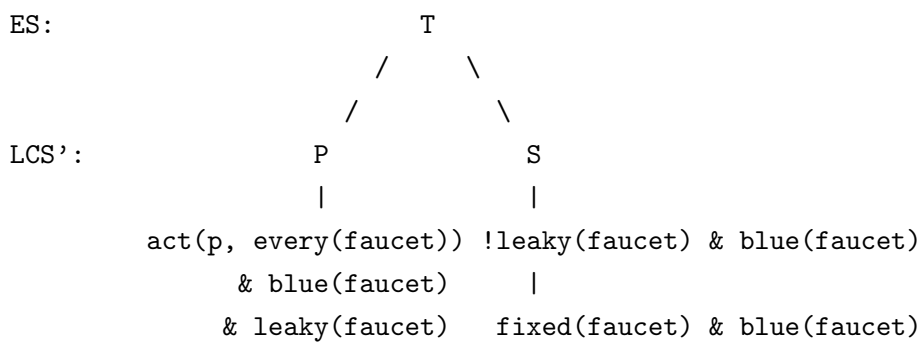
	\begin{verbatim}
	ES:                        T
	                        /     \
	                      /         \
	LCS':               P             S
	                    |             |
	         act(p, every(faucet)) !leaky(faucet) & blue(faucet)
	               & blue(faucet)     |
	              & leaky(faucet)   fixed(faucet) & blue(faucet)
	  
	LCS (1): [p CAUSE [BECOME [EVERY [faucet <fixed>]]]] & [IS [faucet <blue>]]
	LCS (2): [p CAUSE [BECOME [faucets <fixed>]]] & [ARE [faucets <blue>]]
	\end{verbatim}
	{\ehrule}

	NOTE: In the above, I (perhaps astonishingly incorrectly) use EVERY as if it were
	a predicate to enumerate all the faucet instances that could be replaced
	by a normal (``for-all'') symbol (forgot how to spell it in \LaTeX) and
	a variable as we do in predicate logic, or as in LCS (2) use plural form
	instead (this actually helps understand and encode semantics of an NP). LCS (2),
	however, might not indicate the ``for-all'' meaning.
	
\item
	{\bf Mary fixed the flat tire.}

	This entails that the deflated tire became
	full of air and is no longer flat as cause
	by Mary's actions. This would be an achievement.

	\clearpage
	
	{\shrule}
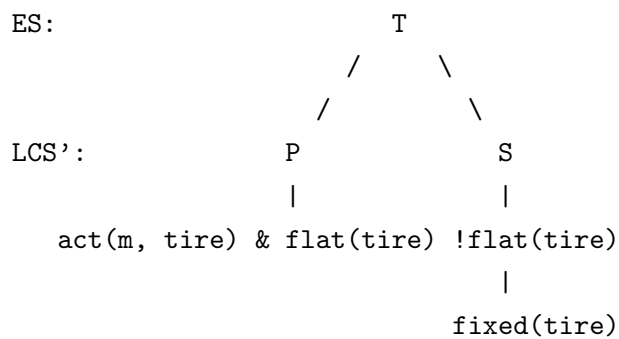
	\begin{verbatim}
	ES:                      T
	                      /     \
	                    /         \
	LCS':             P             S
	                  |             |
	   act(m, tire) & flat(tire) !flat(tire)
	                                |
	                             fixed(tire)
	  
	LCS: [m CAUSE [BECOME [tire <fixed>]]]
	\end{verbatim}
	{\ehrule}

\item
	{\bf John mixed the powdered milk into the water.}

	This indicates that the milk made it to the water
	and the water became {\it milky} vs. {\it milkless} in a sense that
	the two substances fused (regardless the quantity), hence the change of state.
	Likewise, one can argue that the milk became {\it watery}
	vs. {\it dry} (as in powder). Also, since there is an actor, 
	it is an achievement according to \cite{pustejovsky}.

	{\shrule}
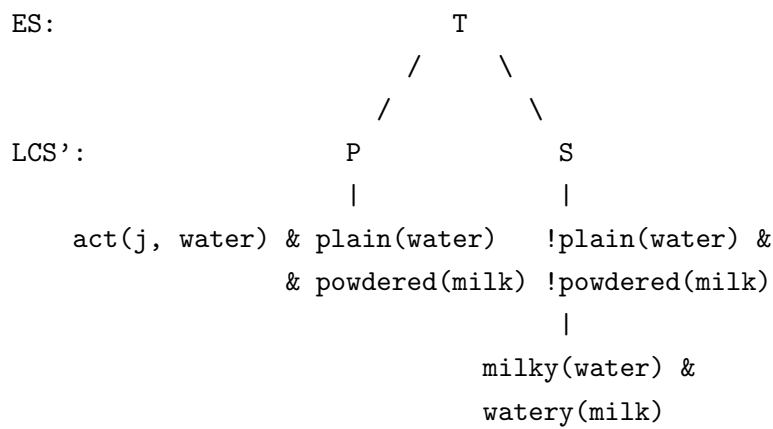
	\begin{verbatim}
	ES:                          T
	                          /     \
	                        /         \
	LCS':                 P             S
	                      |             |
	    act(j, water) & plain(water)   !plain(water) & 
	                  & powdered(milk) !powdered(milk)
	                                    |
	                               milky(water) &
	                               watery(milk)

	LCS: [j CAUSE [BECOME [SUBSTANCE-OF [water <milky>] & [milk <watery>]]]]
	\end{verbatim}
	{\ehrule}

\item
	{\bf The father comforted the crying child.}

	The act of comforting does not necessarily
	imply the child is no longer crying, so there is no
	state transition, but rather a process.

	\clearpage
	{\shrule}
	\begin{verbatim}
	ES:                      P
	                        / \
	                       /   \
	LCS':                e1.....en
	                         |
	      act(f, child) & crying(child) & comfort(child)
	  
	LCS: [f [COMFORT [child <crying>]]] & [f CAUSE [BECOME [child <comforted>]]]]
	\end{verbatim}
	{\ehrule}

\item
	{\bf John painted the white house blue.}

	This is an accomplishment since the house that has
	previously been white became blue in color, so there
	is event culmination.

	{\shrule}
	\begin{verbatim}
	ES:                      T
	                      /     \
	                    /         \
	LCS':             P             S
	                  |             |
	 act(j, house) & house(white) !house(white)
	                                |
	                               house(blue)
	  
	LCS: [j CAUSE [BECOME [house <blue>]]]
	\end{verbatim}
	{\ehrule}

	\clearpage
	
\item
	{\bf Mary rescued the drowning man.}

	This is an accomplishment since the man is no
	longer drowning and it's an agentive action.

	{\shrule}
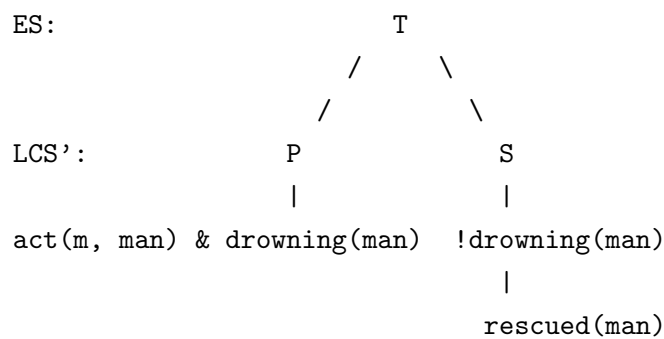
	\begin{verbatim}
	ES:                      T
	                      /     \
	                    /         \
	LCS':             P             S
	                  |             |
	act(m, man) & drowning(man)  !drowning(man)
	                                |
	                               rescued(man)
	  
	LCS: [m CAUSE [BECOME [man <rescued>]]]
	\end{verbatim}
	{\ehrule}

\item
	{\bf Mary cleaned the dirty table.}

	This is a process since the table is not
	necessarily clean after Mary has finished
	the cleaning process.

	{\shrule}
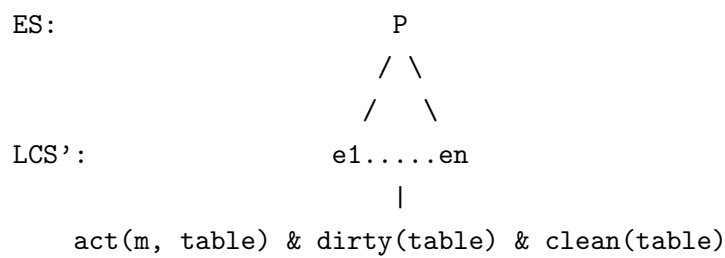
	\begin{verbatim}
	ES:                      P
	                        / \
	                       /   \
	LCS':                e1.....en
	                         |
	    act(m, table) & dirty(table) & clean(table)
	  
	LCS: [m CLEAN [table <dirty>]] & [m CAUSE [BECOME [table <cleaned>]]]]
	\end{verbatim}
	{\ehrule}

	Note, as usual, cleaned table does not mean it is entirely clean, hence
	the sentence does not convey culmination of the cleaning event.

\item
	{\bf The waiter filled every empty glass with water.}

	As a result of waiter's actions all empty glasses have
	changed their state from {\it empty} to {\it full with water}.
	The {\it with water} part, the PP, is an extra bit of information
	indicating the exact type of liquid used to fill the
	glasses with, but does not affect or shift the event type
	in any way, just like (1) or (2). The act of filling for
	the waiter is a process, whereas for emty glasses having
	become full is a change of state, hence a transition, with
	a subtype of accomplishment because the ajective {\it every} 
	makes the completed.

	\clearpage
	
	{\shrule}
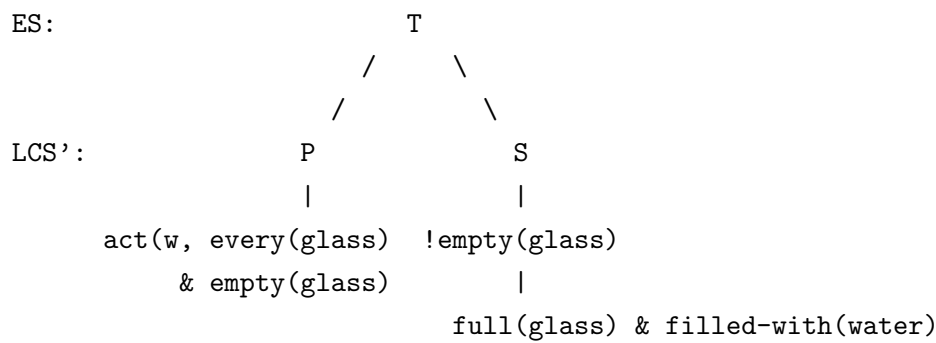
	\begin{verbatim}
	ES:                       T
	                       /     \
	                     /         \
	LCS':              P             S
	                   |             |
	      act(w, every(glass)  !empty(glass)
	           & empty(glass)        |
	                             full(glass) & filled-with(water)
	  
	LCS (1): [w CAUSE [BECOME [EVERY [glass <filled-with-water>]]]]
	\end{verbatim}
	{\ehrule}

	NOTE: the state \verb+<filled-with-water>+ may be broken into
	a finer granularity in LCS predicates like it's done in LCS'.
	
%10
\item
	\begin{enumerate}
	\item
	{\bf John brushed the dirty carpet.}
	\item
	{\bf John brushed the dirty carpet {\it clean}.}
	\end{enumerate}

	These two items are, in fact, quite distinct. (a) is a typical
	process which does not entail the final state of the
	carpet (it might still remain partially dirty), whereas
	(b) is an accomplishment event because it has the transition
	with the
	culmination point from a process to state being clean as
	a result of John's action. Clearly, adverbs may shift the event
	type of a verb from one to another.

	{\shrule}
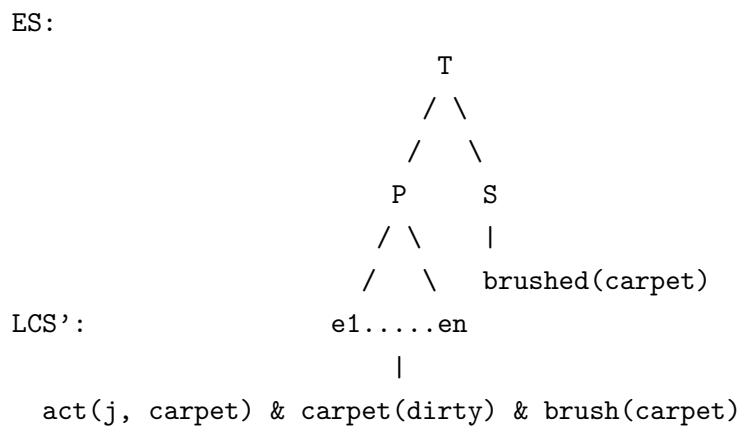
	\begin{verbatim}
	ES:
	                            T
	                           / \
	                          /   \
	                         P     S
	                        / \    |
	                       /   \   brushed(carpet)
	LCS':                e1.....en
	                         |
	  act(j, carpet) & carpet(dirty) & brush(carpet)
	  
	LCS: [j [BRUSH [carpet <dirty>]]] & [j CAUSE [BECOME [carpet <brushed>]]]]
	\end{verbatim}
	{\ehrule}

	There is hidden state change in here of the carpet from {\it not brushed} to
	(somewhat) {\it brushed}, but not necessarily clean 
	(\verb+[carpet <brushed>]+ != \verb+[carpet <clean>]+).
	It occurred to me I have 
	to capture this information.
	
	\clearpage

	The below is for (b):
		
	{\shrule}
	\begin{verbatim}
	ES:                       T
	                       /     \
	                     /         \
	LCS':              P             S
	                   |             |
	act(j, carpet) & carpet(dirty) !dirty(carpet)
	                                 |
	                                clean(carpet)
	  
	LCS: [j CAUSE [BECOME [carpet <clean>]]]
	\end{verbatim}
	{\ehrule}

\end{enumerate}

\section{Lexicon}

First, let us start off with the lexicon of the given example sentences.
The first type of lexical entries in our lexicon, lexemes, will comply with the
feature structure presented in \cite{mokhov-ppt}, \xf{fig:lexeme-orig}. 
The structure was originally derived from \cite{jurafsky} and \cite{arnold}.
For this work
we simplify the lexeme structure by leaving out the phonological form
of the lexeme because it is unused throughout the paper. Likewise, our
lexicon is rather short and we do not deal with polysemy, the SENSE feature
will be a scalar value rather than a set of related senses. 
Hence, our lexeme will have the structure presented \xf{fig:lexeme-simplified}.

The new lexeme representation is a 3-tuple of the form: $name\{X,S,T\}$, where $X$
is the word's spelling, $S$ is its sense, $T$ is its POS tag, and
$name$ is a name of the lexeme, an index, uniquely identifying
the lexical entry in the lexicon. The tag $T$ is of a Penn Tagset, 
\cite{penntagset}. The word sense has been adapted from the online
Webster's dictionary, \cite{webster}.

Pustejovsky and Fong seem to drift away from the feature-based approach
in their work as being not scalable for a decent computational
lexicon. Yet, some feature-based work is preserved, so I will keep
the lexeme as a set of features in addition to new types of lexical
entries presented afterwards.

\begin{figure}
\hrule
\vspace{6pt}
$lexeme$\\
\{

\tab{20}
SPELLING: $x$

\tab{20}
PRONUNCIATION: $y$

\tab{20}
SENSE: $s : \{s1, s2\}$

\tab{20}
POS: $/tag$

\}
\vspace{6pt}
\hrule
\caption{Original Feature Structure of a Lexeme}
\label{fig:lexeme-orig}
\end{figure}

\begin{figure}
\hrule
\vspace{6pt}
$lexeme$\\
\{

\tab{20}
SPELLING: $x$

\tab{20}
SENSE: $s$

\tab{20}
POS: $/tag$

\}
\vspace{6pt}
\hrule
\caption{Simplified Feature Structure of a Lexeme}
\label{fig:lexeme-simplified}
\end{figure}

\subsection{Lexemes}

\subsubsection{Verbs}

\begin{enumerate}
\item
mend$^1$\{\texttt{to mend}, {\it ``to repair; to fix"}, /VB\}

\item
mend$^2$\{\texttt{mended}, {\it ``repaired; fixed"}, /VBD\}

\item
mend$^3$\{\texttt{mending}, {\it ``repairing; fixing"}, /VBG\}

\item
fix$^3$\{\texttt{fixed}, {\it ``repaired"}, /VBD\}

\item
mix$^1$\{\texttt{mixed}, {\it ``formed by mixing components"}, /VBD\}

\item
comfort$^1$\{\texttt{comforted}, {\it ``eased the grief or trouble of; pleased, calmed down"}, /VBD\}

\item
cry$^1$\{\texttt{crying}, {\it ``shedding tears often noisily"}, /VBG\}

\item
paint$^1$\{\texttt{painted}, {\it ``covered with paint"}, /VBD\}

\item
rescue$^1$\{\texttt{rescued}, {\it ``saved from death"}, /VBD\}

\item
drown$^1$\{\texttt{drowning}, {\it ``becoming pulled under the water or other liquid"}, /VBG\}

\item
clean$^1$\{\texttt{cleaned}, {\it ``rid of dirt"}, /VBD\}

\item
fill$^1$\{\texttt{filled}, {\it ``made full; cancelled emptyness"}, /VBD\}

\item
brush$^1$\{\texttt{brushed}, {\it ``cleaned"}, /VBD\}
\end{enumerate}

\subsubsection{Nouns}

\begin{enumerate}
\item
person-cathie\{\texttt{Cathie}, {\it ``first name of a person; agent"}, /NNP\}

\item
person-mary\{\texttt{Mary}, {\it ``first name of a person; agent"}, /NNP\}

\item
person-john\{\texttt{John}, {\it ``first name of a person; agent"}, /NNP\}

\item
dress$^1$\{\texttt{dress}, {\it ``piece of women's clothing; long"}, /NN\}

\item
plumber$^1$\{\texttt{plumber}, {\it ``one who installs, repairs, and maintains piping"}, /NN\}

\item
faucet$^1$\{\texttt{faucet}, {\it ``a fixture for drawing or regulating the flow of liquid especially from a pipe"}, /NN\}

\item
tire$^2$\{\texttt{tire}, {\it ``a rubber cushion that fits around a wheel (as of an automobile) and usually contains compressed air"}, /JJ\}

\item
milk$^1$\{\texttt{milk}, {\it ``a fluid secreted by the mammary glands of females for the nourishment of their young; especially : cow's milk used as a food by humans"}, /NN\}

\item
water$^1$\{\texttt{water}, {\it ``the liquid that descends from the clouds as rain, forms streams, lakes, and seas, and is a
major constituent of all living matter; drinkable consumable''}, /NN\}

\item
father$^1$\{\texttt{father}, {\it ``a man who has begotten a child; agent"}, /NN\}

\item
child$^1$\{\texttt{child}, {\it ``a son or daughter of human parents ; descendant"}, /NN\}

\item
house$^1$\{\texttt{house}, {\it ``a place to live in"}, /NN\}

\item
man$^1$\{\texttt{man}, {\it ``a human being of male gender"}, /NN\}

\item
table$^1$\{\texttt{table}, {\it ``a piece of furniture consisting of a smooth flat slab fixed on legs"}, /NN\}

\item
waiter$^1$\{\texttt{waiter}, {\it ``a person who waits tables (as in a restaurant)"}, /NN\}

\item
glass$^1$\{\texttt{glass}, {\it ``a container made of glass"}, /NN\}

\item
carpet$^1$\{\texttt{carpet}, {\it ``a heavy often tufted fabric used as a floor covering"}, /NN\}
\end{enumerate}

\subsubsection{Adverbs}

\begin{enumerate}
\item
blue$^2$\{\texttt{blue}, {\it ``made being of the color blue"}, /RB\}

\item
clean$^2$\{\texttt{clean}, {\it `made clean; i.e. made not dirty`"}, /RB\}
\end{enumerate}

\subsubsection{Adjectives}

\begin{enumerate}
\item
mended$^1$\{\texttt{mended}, {\it ``repaired; fixed"}, /JJ\}

\item
torn$^1$\{\texttt{torn}, {\it ``broken, split, ripped, pulled apart, rent"}, /JJ\}

\item
red\{\texttt{red}, {\it ``color of ruby or blood"}, /JJ\}

\item
blue$^1$\{\texttt{blue}, {\it ``of the color blue"}, /JJ\}

\item
white$^1$\{\texttt{blue}, {\it ``of the color white : of the color of new snow or milk"}, /JJ\}

\item
leaky$^1$\{\texttt{leaky}, {\it ``permitting fluid to leak in or out"}, /JJ\}

\item
flat$^1$\{\texttt{flat}, {\it ``lacking air : deflated"}, /JJ\}

\item
powdered$^1$\{\texttt{powdered}, {\it ``dried; made of powder"}, /JJ\}

\item
every$^1$\{\texttt{every}, {\it ``being each individual or part of a group without exception"}, /JJ\}

\item
empty$^1$\{\texttt{empty}, {\it ``not full; containing nothing"}, /JJ\}

\item
dirty$^1$\{\texttt{dirty}, {\it ``not clean"}, /JJ\}
\end{enumerate}

\subsubsection{Others}

\begin{enumerate}
\item
the$^1$\{\texttt{the}, {\it ``definite article"}, /DT\}

\item
into$^1$\{\texttt{into}, {\it ``inside of"}, /IN\}

\item
with$^1$\{\texttt{with}, {\it ``used as a function word to indicate the means, cause, agent, or instrumentality"}, /IN\}
\end{enumerate}

\subsection{Semantic Bits}

This section presents types of lexical items, other than feature-based lexemes,
that capture lexical semantics of the lexemes and their composition via
lexical and otherwise relations.

\subsubsection{Event Types}

For our event types of verbs in the sentences in this lexicon we have only 
transitions $T$ and processes $P$. There is also a $S$ state after transition
that could be reflected on the affected objetcs in the example sentences.
There are two types of $T$'s, achievements and accomplishments, let's name
them explicitly as $T_{achievment}$ and $T_{accomplishment}$. 

Thus our event types in the
lexicon:

\begin{itemize}

\item
$T$ 
\item
$T_{achievment}$
\item
$T_{accomplishment}$
\item
$P$
\item
$S$ 
\end{itemize}

\subsubsection{Event Templates}

Other type of lexical entries in our lexicon are event templates
that can be derived from the Semantic Analysis (\xs{sect:semanalysis}).
These entries can be linked to the appropriate lexemes and their
compositions. Note, the variables in these templates also have restrictions
of what they can be (i.e. which lexemes they can be assigned to). For example,
for causative verbs there has to be an animate agent (eg. {\it Cathie})
or a subject that can have an animate role (eg. as {Microsoft Corporation}
in {\it ``Microsoft Corporation was not afraid of law suits against it.''}).
This is to say that a carpet cannot mend a dress, for example.
Making an analogy to security in computing world, these facts can be added to 
the lexicon
as a data structure similar to an ACL (access-control list) matrix that indicates
which semantic capabilities lexemes posses or do not posses based on their roles.

The below is the list of event templates pertinent to the examples
copied nearly verbatim from the ES, LCS' and LCS analysis. A lexical
entry for the event template would

\begin{enumerate}
\item mended-state1\{$T_{achievment}$, \verb+[x CAUSE [BECOME [y <mended>]]]+\}

\item 
	mended-state2\{$T_{achievment}$, \verb+[x CAUSE [BECOME [y <mended>]]] & [IS [y <red>]]+\}
	
	This can be reduced to mended-state1 if we reply on compositionality
	of mended-state1 and plain-single-state below under NPs. Similar
	reduction to the template list entries can be applied to a few items
	below.
	
\item fixed-state1\{$T_{achievment}$, \verb+[x CAUSE [BECOME [EVERY [y <fixed>]]]]+\}
\item fixed-state2\{$T_{achievment}$, \verb+[p CAUSE [BECOME [EVERY [y <fixed>]]]] & [IS [y <blue>]]+\}
\item fixed-state3\{$T_{achievment}$, \verb+[x CAUSE [BECOME [y <fixed>]]]+\}
\item mixed-state1\{$T_{achievment}$, \verb+[x CAUSE [BECOME [SUBSTANCE-OF [y <milky>] & [z <watery>]]]]+\}
\item comforted-state1\{$P$, \verb+[x [COMFORT [y <crying>]]] & [x CAUSE [BECOME [y <comforted>]]]]+\}
\item blue-state1\{$T_{accomplishment}$, \verb+[x CAUSE [BECOME [y <blue>]]]+\}
\item rescued-state2\{$T_{accomplishment}$, \verb+[x CAUSE [BECOME [y <rescued>]]]+\}
\item cleaned-state1\{$P$, \verb+[x CLEAN [y <dirty>]] & [x CAUSE [BECOME [y <cleaned>]]]]+\}
\item filled-water-state1\{$T_{accomplishment}$, \verb+[x CAUSE [BECOME [EVERY [y <filled-with-water>]]]]+\}
\item brushed-state1\{$P$, \verb+[x [BRUSH [y <dirty>]]] & [x CAUSE [BECOME [y <brushed>]]]]+\}
\item clean-state1\{$T$, \verb+[x CAUSE [BECOME [y <clean>]]]+\}
\end{enumerate}

\subsubsection{Semantic Capture of NPs and Adverbials}

\paragraph{NPs}

Meaning of the NPs can be derived from the thematic roles
in the sentence as presented in Chapter 16, \cite{jurafsky} 
(due to lack of time this area remains unexplored at this moment).
Additionally, we can infer some it from the Semantic Analysis
(\xs{sect:semanalysis}) and make up a template-like structure.

For example,
\verb+[IS [y z]]+ can be used to represent a state of a noun $z$ with
a property of $y$. It can be nested ([IS [IS [dress <red>] <torn>]]) and can be stored in the event
template list. This is roughly equivalent to the below, on 
which semantic restriction is
imposed, however (the set of values $z$ and $y$, or JJ and NN, can take):

\begin{verbatim}
NP -> JJ NP
NP -> NN
\end{verbatim}

Thus, for our little lexicon we could write a few entries for the
simple noun phrases we have:

\begin{enumerate}
\item
plain-single-state: \{$S$, \verb+[IS [y z]]+\}
\item
mixture: \{$S$, \verb+[IS [y z] & IS [a b]]+\}
\item
multi-state: \{$S$, \verb+[IS [IS [x y] z]]+\}
\end{enumerate}

\paragraph{Adverbials}

As it has been shown in the analysis, the adverbs, such as
clean$^2$ and blue$^2$ shift the event type from $P$ to $T$.

[TO BE COMPLETED]

This can be encoded as a set of state rules, where input
would be an /RB lexeme, it's corresponding value in the 
referenced by the LET matrix (see below; sorry for jumping
a little), if it yeilds 1, then change the event type of
such a rule from $P$ to $T$. (Again, a matrix??)

[TO BE COMPLETED]

\subsection{Lexicon Analysis}

Compositional and generative properties of our lexicon
would let it scale well as far as lexical semantic concerned.
Using the event templates along with the lexeme's features
we could cover a lot more semantically correct phrases and
sentences without unnecessary lexicon bloat as with pure
feature-based model. This makes the lexicon match the
infinite possibilities of a natural language in capturing
semantics rather than defining it as a finite set of lexemes.

Adding a ``lexeme semantic capabilities'' list (LSCL) could also
restrain the templates the lexemes can be plugged into
w/o cluttering the list of lexemes with pointer to allowed
templates or the reverse. It can also be a lexeme-event-template (LET)
matrix with rows as event templates indexes and columns
are lexemes with a boolean entry indicating which $x$, $y$, and 
$z$ (lexeme variables) allowed to be used in which templates.

Also, another observation that I made is when I put down
all the event templates in the lexicon, with the names that
all had ``-state'' in them. I did not do it intentionally
rather more subconciously. After noticing this, I thought
that {\it possibly} we can model all this template structure
as states only and allowed transitions between them. (This
is a strongly personal opinion. I have not looked at the
FrameNet project myself; only the presentation. Maybe the do the
same?).

Thus a new definition, explicitly stated, of a computational lexicon would be:

\begin{enumerate}
\item
A set of lexemes (feature-based).
\item
A set of event templates.
\item
Either LSCL or LET data structures.
\item
Semantic relation matrix (SRM) among lexemes, as 
in the examples given by Fong in \cite{fongdress} via
antonym/synonym relationship one can find the semantic
opposition and derive the event type based on that.
\end{enumerate}

\section{Summary, Critique, and Conclusions}

The \cite{fongdress} paper mostly focuses on semantic oppositions 
(eg. {\it torn} as opposed to {\it mended} in 1a when related
to dress' final state),
but only briefly touches other types of relations 
(eg. {\it red} vs. {\it torn} when related to dress). Both papers
mostly focus on the lexical semantics of verbs as related to events
and almost no credit given to the semantics of NPs and others 
in relationship to the work done. None explicitly define what
their lexical items in the lexicon would be and why. The presented
material, however, allows to derive the meaning representation
of the noun phrases, adverbials, and prepositional phrases and
how they can shift event type from one another.

It is my belief that the resulting lexicon is quite comprehensive.
The presented instances of it, however, are not perfect and incomplete
and there's no time left to fix them as this is being written. Thus,
I will briefly summarize the current shortcomings of the above lexicon
instance: the lexemes have the sense feature which is currently presented
as just a dictionary (or my own) definition used for the sense in the
example sentences. It has no event types or links to the templates in them.
Instead, I could remove that feature altogether from the lexeme structure
in favour template approach, but keep the rest. I have not provided a concrete
example of the LCSL or LET structures to restrict lexemes to the templates.
Additionally, these structures may suffer from large data sparseness as in
the example of LET matrix making computational and storage aspect
questionable w/o any optimization steps. I have not explored the semantic relations
issue and the necessity of the SRM matrix, which is also likely to be sparse.

Derivation of the new entries for a general-purpose computational
lexicon could be done through the semantic relations and analysis of
verbs argument structure, categories, PPs, which are not explored
in this work.

If this all is not at all sensible and hurts the beautiful mind,
please forgive me for the pain I caused with my work. A lot more
research, information-digesting, and summarizing time is required
to produce a more quality work. Thus, consider this as a humble draft.

%%%%%%%%%%%%%%%%%%%%%%%%%%%%%%%%%%%%%%%%%%%%%%%%%%%%%%%%%%%%%%%%%%

\label{sect:bib}
\bibliographystyle{alpha}
\bibliography{event-structure-torn-dress}

\end{document}